\documentclass[10pt,twocolumn,letterpaper]{article}

\usepackage[pagenumbers]{iccv} % To force page numbers, e.g. for an arXiv version

\usepackage{adjustbox}

\usepackage{multirow}

\usepackage{tabularx}
\usepackage{amsmath} 
\usepackage{float} 
\usepackage{algorithm}

\usepackage[noend]{algpseudocode}
\usepackage{algorithmicx}

\definecolor{iccvblue}{rgb}{0.21,0.49,0.74}
\usepackage[pagebackref,breaklinks,colorlinks,allcolors=iccvblue]{hyperref}

\def\confName{ICCV}
\def\confYear{2025}

\title{Breaking the Box: Enhancing Remote Sensing Image Segmentation with Freehand Sketches}

\author{Ying Zang$^1$, Yuncan Gao$^1$, Jiangi Zhang$^1$, Yuangi Hu$^1$, Runlong Cao$^1$, Lanyun Zhu$^2$,\\ Qi Zhu$^3$, Deyi Ji$^3$, Renjun Xu$^4$, Tianrun Chen$^{4,5}$\thanks{Corresponding Author}\\
$^1$Huzhou University
$^2$Singapore University of Technology and Design\\
$^3$University of Science and Technology of China
$^4$Zhejiang University\\
$^5$KOKONI 3D, Moxin (Huzhou) Tech. Co., LTD.\\
{\tt\small tianrun.chen@kokoni3d.com}
}

\begin{document}
\maketitle
\begin{abstract}

This work advances zero-shot interactive segmentation for remote sensing imagery through three key contributions. First, we propose a novel sketch-based prompting method, enabling users to intuitively outline objects, surpassing traditional point or box prompts. Second, we introduce LTL-Sensing, the first dataset pairing human sketches with remote sensing imagery, setting a benchmark for future research. Third, we present LTL-Net, a model featuring a multi-input prompting transport module tailored for freehand sketches. Extensive experiments show our approach significantly improves segmentation accuracy and robustness over state-of-the-art methods like SAM, fostering more intuitive human-AI collaboration in remote sensing analysis and enhancing its applications.

\end{abstract}
 
\section{Introduction}
\label{sec:intro}

Remote sensing image segmentation plays a crucial role across a wide range of applications, from land use analysis \cite{mohanrajan2020survey} and search and rescue operations \cite{wang2018deep} to environmental monitoring \cite{yuan2020deep}, military intelligence generation \cite{zhang2022progress}, agricultural production \cite{weiss2020remote}, and urban planning \cite{wellmann2020remote}. Accurately extracting desired objects from remote sensing imagery is not just a technical challenge—it is a fundamental enabler of informed decision-making, real-time interventions, and a deeper understanding of our planet's dynamic landscapes.

\begin{figure}[htbp]
    \centering
    \includegraphics[width=0.82\columnwidth]{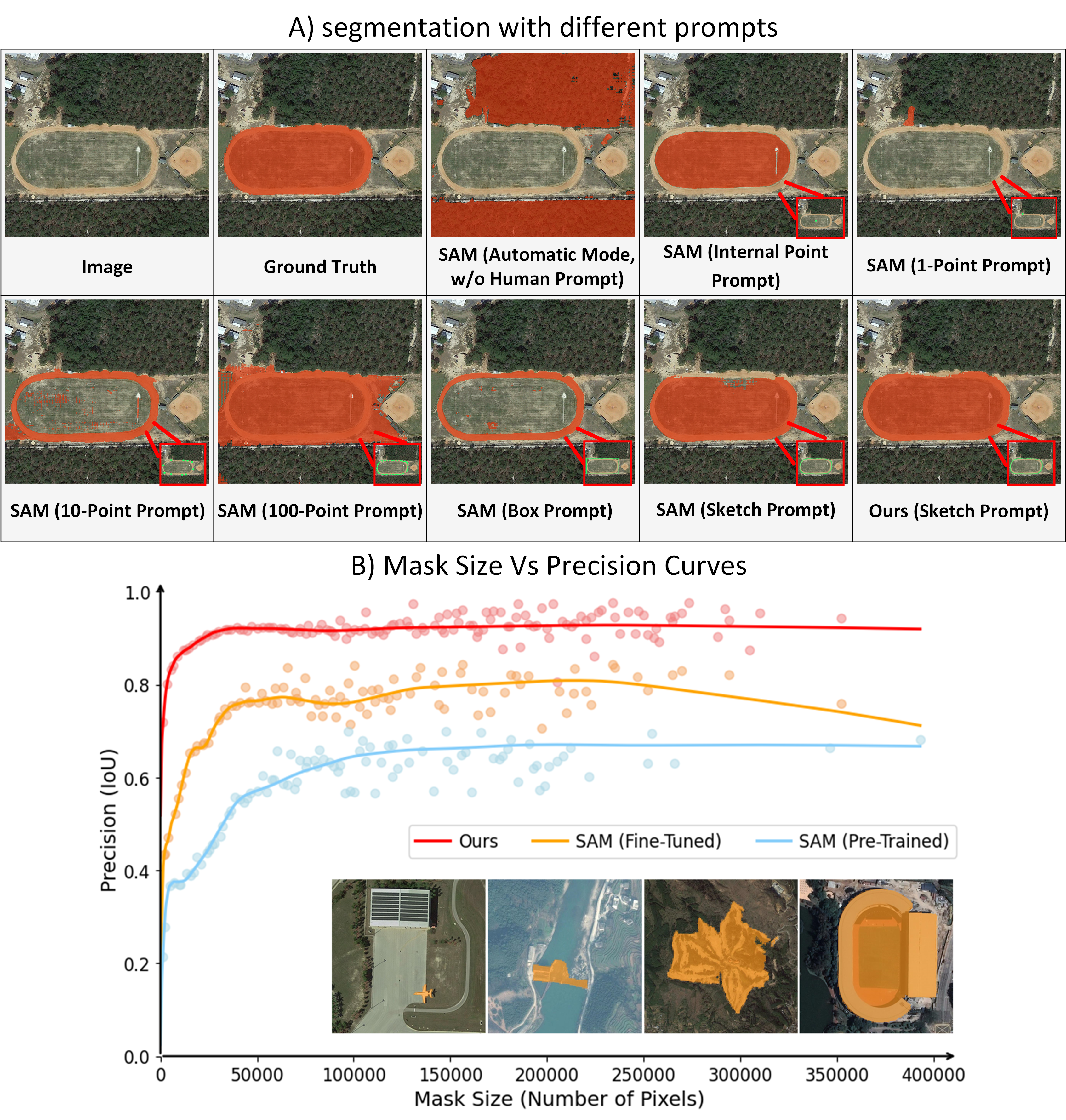}
        \caption  {In this paper, we propose using human freehand sketches (drawing a rough contour of the object) to improve image segmentation in remote sensing images A) Comparison of SAM and our method with different inputs. While SAM struggles with point and box prompts, sketch input improves performance. Red: Segmentation Mask; Green: Prompt; Zoom in for better view; B) Our carefully designed LTL-Net is capable of producing more accurate segmentation masks in all mask size ranges with sketch input compared to vanilla SAM (Pre-Trained) and SAM fine-tuned on remote sensing image dataset with same input prompts, demoted as SAM (Fine-Tuned). More details in Supplementary Material.}
    \label{fig:1}

\end{figure}

However, remote sensing image segmentation presents a unique set of challenges distinct from traditional image segmentation. Unlike everyday objects, which follow predictable scales and orientations, objects in remote sensing images exhibit extreme scale variations—ranging from minuscule targets to vast landscapes. Furthermore, the aerial perspective means that these objects lack the consistent viewpoints seen in natural images, appearing at arbitrary rotations and unpredictable spatial distributions \cite{liu2024rotated}. These factors make segmentation inherently difficult, even for state-of-the-art foundation models trained on large-scale datasets, such as SAM \cite{kirillov2023segment} (Fig. \ref{fig:1} A). Even with strong generalization abilities, these models struggle to accurately segment highly complex and irregularly shaped targets, even when provided with clear prompts. Achieving precise segmentation for arbitrary objects requires effective input prompts. However, through extensive experiments and analysis, we have found that existing prompting methods still have notable limitations. Currently, the most common prompting mechanisms for SAM—points and bounding boxes—are often inadequate for precise segmentation. A single point prompt may fail to capture the full extent of the desired region, while selecting multiple points can inadvertently include extraneous areas, making the process time-consuming and imprecise. Similarly, bounding boxes, while seemingly intuitive, are ill-suited for objects with intricate structures or varying textures, often leading to the inclusion of irrelevant regions or the omission of crucial details (visualized examples in Fig. \ref{fig:1} A and Fig. \ref{fig:sam}).

% Given these challenges—particularly in the zero-shot, interactive segmentation setting of models like SAM—achieving accurate segmentation in remote sensing imagery requires a more robust and flexible approach. To address this, in this work, we propose three key innovations: \textit{\textbf{(1)}} \textit{\textbf{Sketch}} as \textit{\textbf{a novel input prompting approach}} for Interactive Segmentation in Remote Sensing Images, \textit{\textbf{(2)}} LTL-Sensing (Lines That Locate, LTL) as \textit{\textbf{the first dataset}} in the Field with real human drawings of remote sensed imaginary. and \textit{\textbf{(3)}} LTL-Net as \textit{\textbf{a novel network}} with Effective \textit{\textbf{Sketch}} Utilization designed to maximize the benefits of \textit{\textbf{sketch-based}} prompting.
Given these challenges—particularly in the zero-shot, interactive segmentation setting of models like SAM—achieving accurate segmentation in remote sensing imagery requires a more robust and flexible approach. To address this, in this work, we propose three key innovations: {(1)} Sketch as a novel input prompting approach for Interactive Segmentation in Remote Sensing Images, {(2)} LTL-Sensing (Lines That Locate, LTL) as the first dataset in the Field with real human drawings of remote sensed imaginary. and {(3)} LTL-Net as a novel network with Effective Sketch Utilization designed to maximize the benefits of sketch-based prompting.

\noindent\textbf{A Novel Prompting Technique: Freehand Sketch.} Compared to traditional bounding box annotations, we propose that freehand sketch can provide with a more flexible and intuitive way to outline object contours without significantly increasing annotation time. Sketching is one of the most natural ways humans express ideas—everyone, regardless of artistic skill, can draw simple shapes to convey concepts \cite{sayim2011line, hertzmann2020line, goodwin2007isophote, zang2025let, chen2023reality3dsketch, zang2023deep3dsketch+1, chen2024deep3dsketch, zang2024magic3dsketch, zang2023deep3dsketch+, chen2024img2cad, chen2024rapid, chen2024new, chen2023deep3dsketch}. This makes sketches an accessible and user-friendly method for guiding segmentation models. In our approach, a freehand sketch differs from pixel-by-pixel annotation—we do not require users to carefully trace the exact object boundaries. Instead, users can quickly draw an approximate shape that encompasses most of the target region, even if part of it extends beyond the object’s actual boundaries. A rough circular shape is also allowed. Of course, as our experiments confirms, the more precise the sketch, the better the segmentation results (which makes intuitive sense). This allows users to control how much precision they wish to invest in the annotation process.

\noindent\textbf{The First Dataset: LTL-Sensing.} To our knowledge, no existing dataset combines human freehand sketches with images and ground truth segmentation masks in remote sensing. We create drawings for 3,481 remote sensing images, forming the LTL-Sensing dataset. 

\noindent\textbf{The Novel Network: LTL-Net.} Effectively utilizing sketch input in the segmentation process is not straightforward. If not handled properly, sketches can even degrade performance rather than enhance it. Since SAM doesn’t support sketches directly, a naive approach is to sample points along the sketch’s edges and feed them into SAM’s (point) prompt encoder (In theory, sampling all edge points should capture the entire sketch contour). However, freehand sketches are not perfect object outlines—they are meant to be quick and approximate. As a result, blindly using all sampled points can introduce unwanted regions into the segmentation. Another approach is to encode the entire sketch and input it directly into SAM prompt encoder. However, sketches differ fundamentally from synthetic edge maps—different users draw sketches in varying styles, and every sketch contains imprecision. Addressing these inconsistencies and errors requires a dedicated architecture. Therefore, we propose a novel network, LTL-Net, designed to fully harness the potential of sketch-based prompting. LTL-Net features a sketch augmentation module, which generates diverse sketch variations using Bézier curve fitting and controlled perturbations. These augmented sketches are then processed through a carefully designed multi-input prompting mechanism, leveraging optimal transport theory to effectively map mask pixels to multiple sketch inputs. This approach inherently improves robustness to variations in user sketches. Finally, unlike SAM that directly encodes features into a prompt encoder, we introduce a customized image-sketch feature fusion mechanism before SAM’s prompt decoder. Extensive experiments show LTL-Net outperforms existing methods, achieving state-of-the-art results (in all target size range, as shown in Fig. \ref{fig:1} B).

By introducing a novel sketch-based prompting approach, the LTL-Sensing dataset, and the new LTL-Net, we believe that this work not only improve segmentation accuracy and efficiency but also enhance our understanding of how humans interact with remote sensing imagery, which ultimately holds potential in contributing to better decision-making in a variety of fields on Earth.

\section{Related Work}
\label{sec:relate}

\noindent\textbf{Interactive Segmentation and Segment Anything in Remote Sensing:}
Interactive segmentation allows users to define and refine image segmentation results through collaborative interactions with algorithms \cite{kitrungrotsakul2020interactive,wei2023focused}. The most common approach for human-computer interaction in segmentation the interactive segmentation setting is through mouse clicks \cite{chen2022focalclick,jang2019interactive,zhu2023continual,sofiiuk2020f,sofiiuk2022reviving, zhu2021learning,zhu2024addressing}. Recently, this field is revolutionized by models like Segment Anything (SAM) and its variants, which are trained on a vast amount of data and are capable of segmenting objects based on points or box input
 with impressive accuracy in a zero-shot manner. SAM has already been used in the field of remote sensing to reducing the need for manual annotation or create new datasets \cite{ma2024sam, osco2023segment, zhang2024geoscience, sultan2023geosam, luo2024sam, ma2024manet, qiao2025sam}. We hereby propose both new prompt input and new network configurations to elevate the performance for SAM in remote sensing.

\noindent\textbf{Referring Image Segmentation in Remote Sensing and Sketch-Guided Segmentation:}
Another field that is close to this research is referring image segmentation, in which user ``say" what they want to segment \cite{hu2016segmentation,li2018referring, zhu2024llafs, luo2020cascade,hu2020bi,zang2025resmatch}. Such the task has also been investigated in remote sensing domains \cite{sun2022visual, zhan2023rsvg, chen2025rsrefseg, zhang2025referring, li2025scale, dong2024cross}, but it is not hard to find that the text input is sometimes insufficient to express ``what user want to segment" . Another limitation is that text input has minimal spatial cues and there is a huge domain gap between textual input with the image domain \cite{liu2024rotated}. 

Here, we explore an alternative input format—freehand sketches for users to ``see it, sketch it, segment it". Sketches are used to guide image retrieval for decades \cite{del1997visual, dey2019doodle, eitz2010sketch, koley2024handle, matsui2017sketch, qi2016sketch, hou2011detection}. Sketches have recently been introduced to (zero-shot) object detection \cite{chowdhury2023can}. But very few researches have investigated sketch-guided segmentation. The comparison  between our method and some key methods is shown in  Table. \ref{Comparative}. Unlike previous approaches that treated sketches as mere input prompts—essentially substituting text or fixed class labels \cite{hu2020sketch}—we encourage users to draw a sketch in the area they wish to segment (with spatial specification) allowing human knowledge to enhance segmentation performance (with human-guided improvement on performance).

\begin{table}[htbp]
\caption{Comparison between our approach and some similar sketch-enabled and interactive segmentation approach}
\centering
\renewcommand{\arraystretch}{1.8} % 调整表格行高，增加垂直间距
\resizebox{\columnwidth}{!}{ % 自动调整表格宽度以适应单栏
\begin{tabular}{|c|c|c|c|c|}
\hline
\multicolumn{1}{|c|}{Method} & \parbox{3cm}{Sketch-Assisted Saliency Detection \cite{bhunia2023sketch2saliency}} & \parbox{3cm}{Sketch-based Segmentation \cite{hu2020sketch}} & \parbox{3cm}{Interactive Segmentation Foundation Model \cite{kirillov2023segment}} & Ours \\ \hline
Spatial Specification &  &  & \checkmark & \checkmark \\ \hline
Human-guided Improvement & \checkmark & \checkmark & \checkmark & \checkmark \\ \hline

Input Error-Tolerant Flexibility  &  & \checkmark &  & \checkmark  \\ \hline
\end{tabular}
}
\vskip -0.1in
\label{Comparative}
\end{table}

\section{Drawing Freehand Sketch–A Pilot Study and a New LTL-Sensing Dataset}

% \begin{table}[H]
% \caption{---}
% \vskip -0.25in
% \label{sample-table}
% \begin{center}
% \resizebox{\columnwidth}{!}{ % 调整表格宽度以适应栏宽
% \begin{small}
% % \begin{sc}
% \begin{tabular}{c|ccccccc}
% \toprule
% Model  & P@0.5 & P@0.6  & P@0.7 & P@0.8 & P@0.9  & oIoU  & mIoU  \\
% \midrule
% SAM  &91.78    &82.02   &68.28  &51.25   &28.41 & 78.17 & 76.31 \\
% SAM HQ &85.78  & 75.24   &62.77 &48.23 &26.46  & 81.23  & 73.64\\
% % $\times$    &$\surd$   &55.62 &36.00  &9.94 &63.13 &48.53\\    \\
% \bottomrule
% \end{tabular}
% % \end{sc}
% \end{small}
% }
% \end{center}
% \vskip -0.1in
% \label{tab:3}
% \end{table}

As previously mentioned, remote sensing objects significantly differ from typical images, as they contain highly diverse information with varying scales and orientations. When users seek to specify the segmentation of any object in remote sensing images, it is crucial to guide the process with appropriate input information. Traditional interactive segmentation methods that utilize points or boxes as inputs can lead to extraneous or incomplete results. For instance, in the second example in Fig. \ref{fig:1} B), the dam displays a complex shape with features connected to surrounding objects, such as the riverbank, making it challenging for point inputs to accurately identify the boundaries of these connected areas. Similarly, the sports stadium on the right side of Fig. \ref{fig:1} B) consists of two distinct regions—the stands and the field—yet users may wish to segment the entire stadium. When using box inputs, the network often struggles to determine whether to segment the entire compound object or just a specific area, as illustrated in Fig \ref{fig:sam}, where the SAM model only segmented portions of the field and the stadium stands.

To address this issue, we propose using sketches, a method accessible to everyone. To validate this approach, we gathered five volunteers and conduct experiments using the existing RRSIS-D dataset \cite{liu2024rotated}, originally designed for text-referred segmentation. Each image in this dataset includes a textual description of the object the user wishes to segment, accurately reflecting the actual "object" rather than arbitrary regions. Notably, the annotators are not required to fit the sketch to the actual contours; whether the sketch is inside or outside the object’s edges is acceptable, as strict adherence would be impractical and time-consuming. 

In practice, we employed custom software on tablets for this task, allowing volunteers to sketch with pencils directly on the images without any ground truth (GT) mask reference, relying solely on the text prompt from the original dataset. To ensure efficiency, we set a drawing limit of 30 seconds for each sketch and timed each session. After each session, another annotator verified the results. Ultimately, the average drawing time was measured at 15 seconds per image after annotating all 3,481 images from the test set of the RRSIS dataset. We found that this time consumption is sufficient to acquire high-quality guidance information, achieving a mean Intersection over Union (IoU) of 12.48\% when compared to the corresponding edges detected by Canny edge detectors. We believe this dataset is not only valuable for this research but also has broader implications for understanding how we human interact with remote sensing images. For more details and visualization of the LTL-Sensing dataset, please refer to the Supplementary Material.

\section{The LTL-Net}
\subsection{Network Backbone and Sketch-Image Fusion}
\begin{figure*}[htbp]
    \centering
    \includegraphics[width=0.88\textwidth]{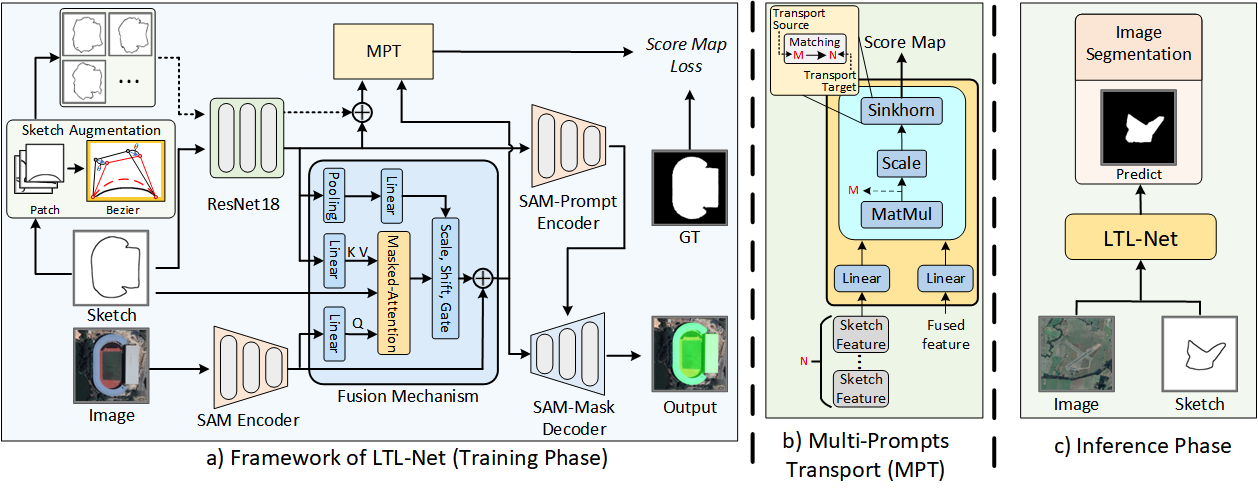}
    \caption{a) Overall structure of LTL-Net.  Masked Attention, Sketch augmentation, and Multi-Prompts Transport are introduced in this sketch-based task and are used in training to get elevated performance. The GT mask is used for supervision. b) we introduce Multi-Prompts Transport (MPT).This cross-prompt coordination mechanism enhances segmentation precision while maintaining consistency across varying sketch inputs. c) Inference Phase: Input image and freehand sketch to output segmentation aligned with sketch guidance.}
    \label{fig:2}
    \vspace{-0.5cm}
\end{figure*}

The network structure is shown in Fig. \ref{fig:2}. We use the SAM model as our backbone network, which consists of an image encoder, a prompt encoder, and a decoder. Images \( I \in \mathbb{R}^{H \times W \times 3} \) are encoded by the encoder to obtain features \( F_I \), which are then passed into the decoder. Meanwhile, input prompts are encoded by the prompt encoder and integrated into the decoding process. To enhance this setup, we aimed to improve the control provided by the sketches, recognizing that sketches can offer more nuanced and intuitive guidance compared to traditional prompt inputs. By leveraging the unique characteristics of sketches, we sought to enable more precise segmentation outcomes that align closely with user intentions. To achieve this, we introduced a Fusion Mechanism before the decoding step, which has been validated as effective in previous works \cite{suzek2007uniref, wu2023uniref++}. The sketches, encoded using ResNet \cite{he2016deep} to produce features \( F_S \), were directed into the SAM decoder while also passing through this Fusion Mechanism. This mechanism allowed us to impose constraints on the feature inputs prior to the decoder.

%During our research on the SAM (Segment Anything Model) network, we observed that it consists of an image encoder, a prompt encoder, and a decoder. The image \(I\in\mathbb{R}^{\mathsf{H}\times\mathsf{W}\times\mathsf{3}}\) is encoded by the image encoder to generate the feature \({F_I}\), which is then fed into the decoder. Meanwhile, the input prompt is encoded by the prompt encoder and integrated into the decoding process. To enhance the regulatory effect of sketches on segmentation results, we replaced the prompt encoder with a sketch encoder, allowing the sketch \(S_K\in\mathbb{R}^{\mathsf{H}\times\mathsf{W}}\) to serve as the prompt input to the model. Experimental results demonstrate that this modification effectively improves model performance (see Table \ref{tab:1}). On this basis, we further introduced a fusion mechanism to optimize the feature inputs and strengthen the regulatory effect of sketches. The detailed

Specifically, the sketch \(S_K\) is encoded by ResNet \cite{he2016deep} to obtain features \(F_S\) following by a fusion mechanism. Image features \(F_I\) are used to obtain Queries (Q), while the keys (K) and values (V) are derived from the sketch features \(F_S\). These inputs, after linear projection, goes through a multi-head cross-attention, following a pooling operation and then regress the scale, shift, and gating parameters \(\gamma, \ \beta, \ \alpha,\). Finally, through residual connections, the output features are added on the original visual features \(F_I\) to enhance control over the decoder. The process is represented as follows:
\begin{equation}
  \gamma, \beta, \alpha = Linear(Pooling(F_S))
  \label{eq:1}
\end{equation}
\vspace{-0.5cm}
\begin{equation}
  F_U = F_I + \alpha \cdot (Attention(Q, K, V) \odot (1 + \gamma) + \beta)
  \label{eq:2}
\end{equation}

Where \(Attention(\cdot) \) denotes the Attention operation, \( \odot \) denotes element-wise multiplication, and \( F_U \) is the final output of the fusion module. Next, we use the SAM-Prompt Encoder to extract features \( F_S \), denoted as \( F^{'}_S \). Finally, the SAM-Mask Decoder uses \( (F_U + F^{'}_S) \) to generate a binary mask for the target object \( O \in \mathbb{R}^{H \times W} \). 

\subsection{Masked Attention}
Our sketches are, in fact, quite sparse; apart from the areas containing strokes, most regions have no information (0). Standard cross-attention distributes attention broadly across spatial locations, leading to target ambiguity when multiple similar objects appear in an image. Self-attention, while preserving local feature structures, fails to effectively incorporate sketch guidance. Instead, we opted for a mask attention approach, applying it to the binarized sketches and the images only focusing on the regions that have strokes information. We generate an attention mask \( M \in \mathbb{R}^{H' \times W'} \) in the following manner:

\begin{equation}
M(x, y) = 
\begin{cases} 
0 & \text{if } S(x, y) = 1 \text(with Stroke) \\
-\infty & \text{otherwise}
\end{cases}
\label{eq:3}
\end{equation}

This mask is introduced into the attention weight calculation, forcing the model to focus on the sketch area, represented as follows:
\begin{equation}
Attention(Q, K, V) = softmax(M + QK^\top)V
\label{eq:4}
\end{equation}

Our masked attention mechanism explicitly constrains the attention computation to sketch-relevant regions, providing improved boundary delineation and enhanced multi-object disambiguation: By focusing only on sketch-relevant areas, the mechanism achieves 3.5\% higher oIoU and 11\% higher IoU than standard attention. These advantages are particularly important for remote sensing applications, where images often contain multiple similar objects (e.g., buildings, vehicles) at various scales and orientations.

\subsection{Sketch Augmentation and Multi-Prompts Transport}
Since our approach relies on freehand sketches, some degree of imprecision is inevitable. Users naturally exhibit diverse drawing styles and varying levels of detail, leading to an infinite number of possible sketches corresponding to the same ground truth (GT) mask. The key challenge here is to harness this variability effectively—capturing the essential features of the sketches while ensuring that these natural inconsistencies do not compromise segmentation performance. To address this challenge, we adopt a two-step approach: (1) First, we design a \textbf{sketch augmentation} method that synthesizes diverse sketches with varying styles and levels of imprecision, mimicking how different users sketch with different levels of detail. This allows the model to learn from a broader range of possible inputs and improves its robustness to real-world variations. (2) Second, we introduce an \textbf{optimal (multiple-prompt) transport mechanism} that enables the network to learn from multiple sketches corresponding to the same GT mask. As sketches contain varying degrees of uncertainty along different segments (e.g., hand tremors may create imprecision in some areas but not others), optimal transport (OT) offers a theoretically sound framework for handling this uncertainty by formulating the problem as distributional matching between sketch and image features. Unlike attention, which normalizes row-wise, OT enforces dual stochastic constraints that balance both feature-to-pixel and pixel-to-feature associations. By leveraging global distribution matching to achieve optimal alignment between multiple prompts and image pixels, our method ensures more precise control over the mask generation process, enhancing segmentation accuracy and consistency. 

\noindent\textbf{Sketch Augmentation.} 
%Existing multi-prompt methods have limitations in segmentation tasks. Traditional approaches align prompts with pixel embeddings through simple cross-attention, which can lead to attention confusion among score maps from different prompts. This makes features indistinguishable and reduces segmentation accuracy. To address this issue, this paper draws on the mathematical framework of Optimal Transport (OT), which explicitly models the optimal alignment between (multiple) prompts and pixels through global distribution matching. Inspired by Sinkformer's use of the Sinkhorn algorithm to replace SoftMax in unimodal tasks, we designed Multi-Prompts Transport (MPT). The dual stochastic constraints of Sinkhorn force each prompt to selectively focus on specific pixel regions while avoiding redundancy or omissions, thereby enhancing the diversity of multiple prompts. MPT achieves a refined alignment between multi-prompts and pixel embeddings through the dynamic allocation mechanism of OT, breaking through the passive alignment bottleneck of traditional attention mechanisms. Then, how to obtain diverse sketch data is a key issue.
Given that having different individuals draw in various styles requires significant time and resources, we adopted a synthetic approach based on Bézier curve fitting and perturbation to simulate this process. Specifically, we employ a Bezier Pivot Based Deformation (BPD) strategy, which generates a variety of sketch variants by adjusting the control points of Bezier curves. We selected Bezier curve fitting for sketch augmentation based on three key considerations: (1) its parametric nature allows controlled perturbation while preserving the essential shape characteristics, (2) its mathematical properties enable smooth interpolation that mimics natural hand movement variations, and (3) its computational efficiency permits rapid augmentation during training. Other alternatives we explored included B-splines and Fourier descriptors, but they either introduced excessive complexity or failed to preserve critical shape features under perturbation.

In practice, we first employ a morphological skeletonization method to extract the centerline of the sketch, which removes the boundary pixels while preserving the topological structure of the image. Given the thickness of the sketch lines, we then extract the centerline before segmentation to avoid splitting the lines into multiple fragments. The processed sketch is then divided into multiple non-overlapping square patch. Within each patch, the largest connected set of pixels is selected as the principal curve and fitted using a cubic Bézier curve, with the formula as follows:
\begin{equation}
f = (1-t)^3 p_0 + 3t(1-t)^2 p_1 + 3t^2(1-t) p_2 + t^3 p_3
\label{eq:5}
\end{equation}

Where \( t \in [0,1] \), \( p_0 \) and \( p_3 \) are the start and end points of the curve. The pivots \( p_1 \) and \( p_2 \) are the control points that determine the shape of the curve. To ensure the fitting effect, curves containing only a very few pixels are discarded. By adjusting the coordinates of the Bezier curve's control points, a series of new curves can be generated, thus creating a variety of variants for each sketch block. We apply a random displacement \( \theta \) to the control points, generating new control point positions \( p' \) by adjusting the original control points \( p \), and the relationship can be expressed as:
\begin{equation}
p' = p + \theta
\label{eq:6}
\end{equation}

After we obtain the parametric representation of the hand-drawn sketch, we apply random displacements to the control points of the Bezier curves. The magnitude of the displacement is dynamically adjusted according to the number of strokes in the sketch:

\begin{equation}
\theta = floor\left(\frac{row}{C}\right) \times K
\label{eq:7}
\end{equation}

\begin{figure}[htbp]
    \centering
    \includegraphics[width=0.8\columnwidth]{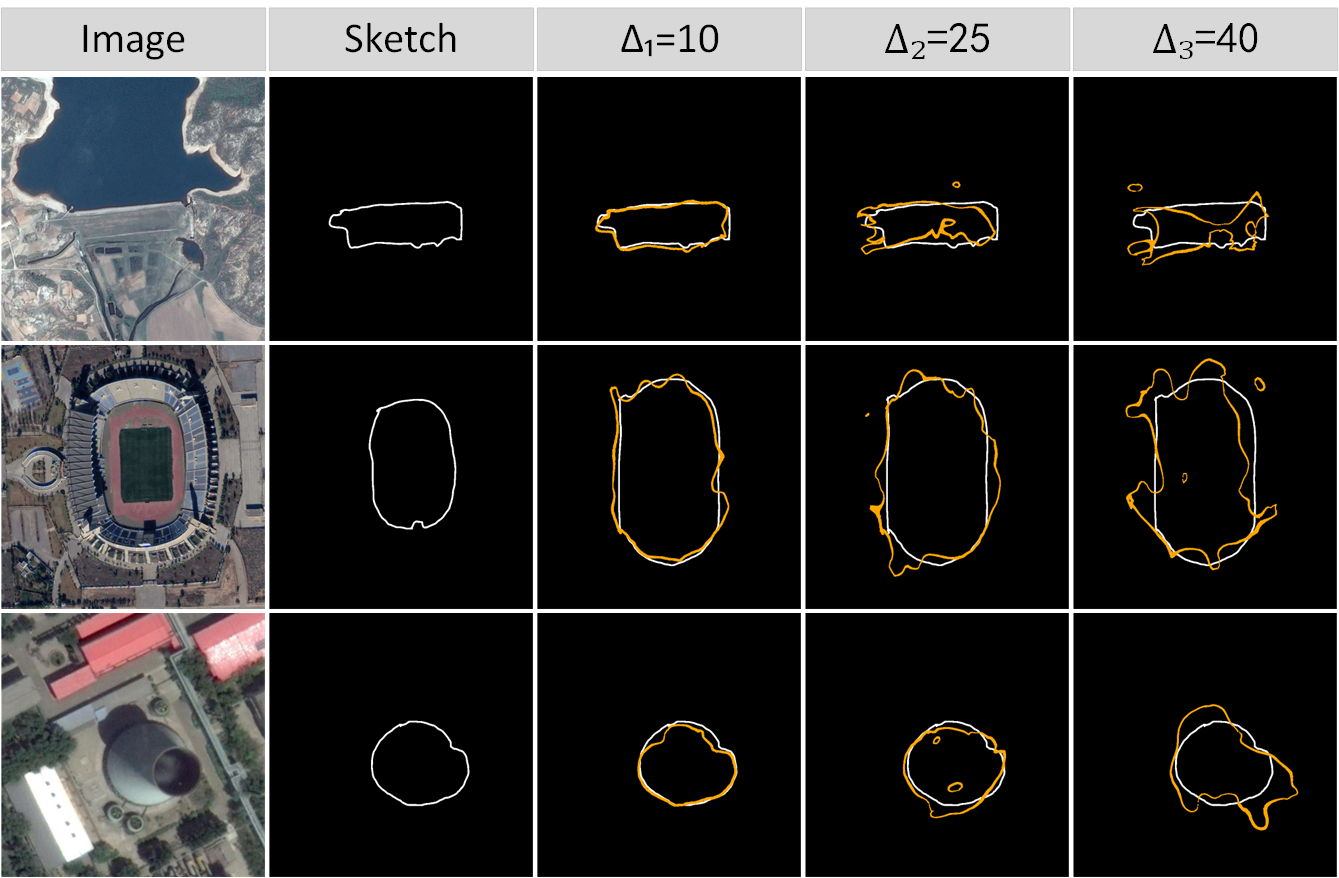}
    \caption{Visualization of sketches augmentation with different amplitude of pertubation.}
    \label{fig:3}

\end{figure}

Where \( C \) represents the control unit for the number of rows in displacement deformation, \( K \) represents the amount by which the displacement magnitude increases for each additional row in the sketch, \( floor(\cdot) \) denotes the floor function, and \( C \) and \( K \) are two key parameters. As the number of rows in the sketch increases, the displacement magnitude gradually increases, ensuring that sketches of different scales undergo corresponding deformations, thus achieving a balance between flexibility and stability in deformation. Visualized examples are shown in Fig. \ref{fig:3}.

\noindent\textbf{Multi-Prompt Transport.}
After generating sketch variants for a single sample, we need to effectively map them to a single image-GT pair when training on each samples. Our goal is to align the sketches (multiple prompts) with extracted image features to ensure robust feature extraction while remaining invariant to individual drawing styles and skill levels. Here, we turn to optimal transport theory to perform the alignment, which is a powerful framework for handling such alignment problems by finding the most efficient mapping between distributions. 

Let us first recall the fundamental goal of optimal transport theory: to quantify and minimize the distributional divergence between two probabilistic measures. Let \( F = \{f_i\}_{i=1}^M \) and \( G = \{g_j\}_{j=1}^N \) represent two sets of feature vectors, the corresponding discrete measures are defined as:
\begin{equation}
\mathbf{u} = \sum_{i=1}^M \mu_i \delta_{f_i}, \quad 
\mathbf{v} = \sum_{j=1}^N \nu_j \delta_{g_j},
\end{equation}
where \(\delta_{(\cdot)}\) denotes Dirac delta functions, and the weight vectors satisfy probabilistic constraints: \(\sum \mu_i = 1\), \(\sum \nu_j = 1\).

The core optimization problem in discrete optimal transport can be mathematically expressed as:
\begin{equation}
\begin{aligned}
\mathbf{T}^* &= \arg\min_{\mathbf{T} \in \mathbb{R}^{M \times N}} \sum_{i,j} \mathbf{T}_{ij} \mathbf{C}_{ij} \\
\text{s.t} \quad & \mathbf{T1}^N = \boldsymbol{\mu}, \quad \mathbf{T^\top1^M}= \boldsymbol{\nu},
\label{eq.new}
\end{aligned}
\end{equation}
where the cost matrix \(\mathbf{C}\) typically reflects pairwise dissimilarities, such as \(\mathbf{C}_{ij} = 1 - \frac{f_i g_j^\top}{\|f_i\|_2 \|g_j\|_2}\) for cosine-based measurements.

As a common practice, entropy-regularized optimal transport can be used to circuvent the computational bottlenecks in solving Eq. \ref{eq.new}:
\begin{equation}
\begin{aligned}
\mathbf{T}^* &= \arg\min_{\mathbf{T} \in \mathbb{R}^{M \times N}} \sum_{i,j} \mathbf{T}_{ij} \mathbf{C}_{ij} - \epsilon H(\mathbf{T}) \\
& \text{s.t} \quad  \mathbf{T1}^N = \boldsymbol{\mu}, \quad \mathbf{T^\top1^M}= \boldsymbol{\nu},
\end{aligned}
\end{equation}
where \(H(\mathbf{T}) = \sum_{i,j} \mathbf{T}_{ij} \log \mathbf{T}_{ij}\) denotes the Shannon entropy. 

To derive a solution, we turn to the Sinkhorn algorithm, which can provide with an iterative solution:
\begin{equation}
\mathbf{T}^* = \text{diag}(\mathbf{a}^t) \exp\left(-\mathbf{C}/\epsilon\right) \text{diag}(\mathbf{b}^t),
\end{equation}
with scaling vectors updated as \(\mathbf{a}^t = \boldsymbol{\mu} \oslash (\exp(-\mathbf{C}/\epsilon) \mathbf{b}^{t-1})\) and \(\mathbf{b}^t = \boldsymbol{\nu} \oslash (\exp(-\mathbf{C}/\epsilon) \mathbf{a}^t)\), initialized with \(\mathbf{b}^0 = \mathbf{1}\). Following \cite{schmitzer2019stabilized}, logarithmic domain computations is used to enhance numerical stability.

Equipped with such the tool, we can calculate an alignment score map \( S \in \mathbb{R}^{HW \times KN} \) based on multiple sketch features \( h_{sk} \in \mathbb{R}^{M \times D} \) and the extracted (fused) image feature \( h_{img} \), where \( M = H \times W \) (the product of the image's height and width), and \( D \) is the embedding dimension:

\begin{equation}
S = h_{sk} h_{img}^\top
\label{eq:8}
\end{equation}

Where  \(h_{img}\) are the (fused) image feature processed by Fusion Mechanism, To transport the distribution of multiple sketches to the pixel distribution, we define a total cost matrix \( C \), which is calculated based on the alignment score map \( S \), specifically \( C = 1 - S \), where \( C \in \mathbb{R}^{HW \times KN} \) represents the cost matrix. Based on this cost matrix \( C \), our goal is to obtain the corresponding optimal transport plan \( T^* \) through the Multi-Sketch Transport (MPT) algorithm, which aims to assign each image pixel to multiple sketch prompts, allowing multiple sketch prompts to be associated with each pixel. Therefore, \( T^* \), as a mapping matrix, maximizes the cosine similarity between multimodal embeddings. Through the MPT algorithm, we obtain the optimized score map as:

\begin{equation}
S^* = M(T^* \odot S)
\label{eq:9}
\end{equation}

Where \( M(\cdot) \) is a reshaping operation, which first reshapes \( \mathbb{R}^{HW \times KN} \) into \( \mathbb{R}^{HW \times K \times N} \), and then sums over the \( N \) dimension for all score maps, the score map \( S^* \in \mathbb{R}^{HW \times K} \), optimized by the optimal transport plan \( T^* \), can serve as independent logical values for segmentation masks. After obtaining the optimized score map \( S^* \), we upsample it to match the size of the original image to get a predicted results of the optimized score map  \( Y \in \mathbb{R}^{H_i \times W_i \times K} \).

We use GT mask as the reference to guide the optimization of the score map, with the loss function as follows:
\begin{equation}
L_{Focal}(Y, T) = \frac{1}{HW} \sum_{i=1}^{H} \sum_{j=1}^{W} \alpha_t \cdot L_{CE}(y_{i,j}, t_{i,j}) \cdot (1 - p_t)^\gamma
\label{eq:focal_loss}
\end{equation}

in which 
\begin{equation}
L_{CE}(y_{i,j}, t_{i,j}) = -[t_{i,j} \log(y_{i,j}) + (1 - t_{i,j}) \log(1 - y_{i,j})]
\label{eq:cross_entropy_loss}
\end{equation}

Where \( y_{i,j} \) and \( t_{i,j} \) are the values of the \( i,j \)-th pixel of \( Y \) and \( T \), respectively. \( \alpha_t \) and \( (1-p_t)^\gamma \) are two balancing factors:

\begin{equation}
p_t = y_{i,j} \cdot t_{i,j} + (1 - y_{i,j}) \cdot (1 - t_{i,j})
\label{eq:pt}
\end{equation}

\begin{equation}
\alpha_t = \alpha \cdot t_{i,j} + (1 - \alpha) \cdot (1 - t_{i,j})
\label{eq:alpha_t}
\end{equation}

%-------------------------------------------------------------------------
\section{Experiments}
\subsection{Implementation Details}
In our experiment, images and GT masks are derived from RRSIS-D dataset \cite{liu2024rotated}, and we also use their train-test split. The model is trained on an NVIDIA A800 Tensor Core GPU. The initial learning rate was set to 2e-4 with a weight decay coefficient of 0.05. We employed a linear warm-up strategy with a warm-up factor of 1.0, completing the warm-up over the first 200 iterations. The learning rate underwent a step decay at the 75,000th iteration, with the maximum number of iterations set to 120,000. According to the aforementioned settings, the entire training process took approximately 15 hours. Following prior works \cite{wu2020phrasecut, zhan2023rsvg}, we utilize Overall Intersection-over-Union (oIoU), Mean Intersection-over-Union (mIoU), and Precision@X (P@X) as evaluation metrics. For more implementation details, please refer to the Supplementary Material.

\subsection{Sketch Boost the Segmentation Performance}
In our experiment, we validate two key claims: (1) sketches serve as a powerful tool for enhancing segmentation performance across various existing models, and (2) our proposed network, LTL-Net, is highly effective—its architecture and training protocol, specifically designed for sketch input, establish a new state-of-the-art approach in this field.

\begin{figure}[htbp]
    \centering
    \includegraphics[width=\columnwidth]{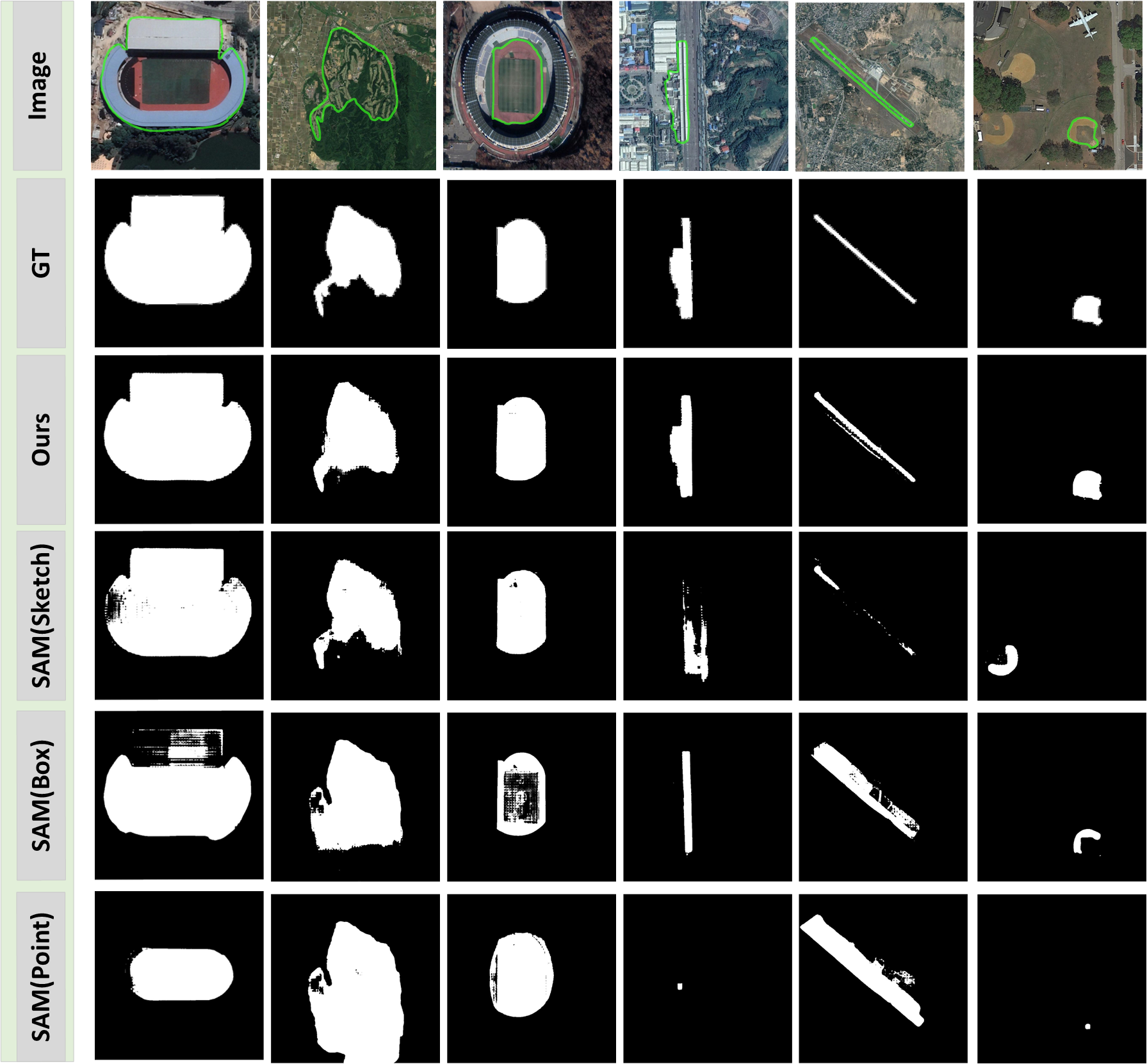}
    \caption{Segmentation Result from Various Input Prompts. Details in Supplementary Material.}
    \label{fig:sam}
    
\end{figure}

First, as shown in Fig. \ref{fig:sam}, incorporating sketch input significantly improves the performance of the Segment Anything Model (SAM), enabling more precise identification of target areas. Compared to the fine-tuned SAM (denoted as SAM*), which utilizes text input from the RRSIS-D dataset, sketch-based input consistently yields superior performance, as evidenced in Tab. \ref{tab:1}.

Second, as demonstrated in Tab. \ref{tab:1}, our method outperforms other referring image segmentation approaches that accept sketches as input. To further assess the impact of sketch guidance, we modified both the original SAM and fine-tuned SAM (SAM*) to accept sketches as input. Our method, specifically designed to leverage sketch-based guidance, achieves state-of-the-art performance. Additional experimental details are provided in the Supplementary Material.

\begin{figure}[htbp]
    \centering
    \vspace{-0.3cm}
    \includegraphics[width=\columnwidth]{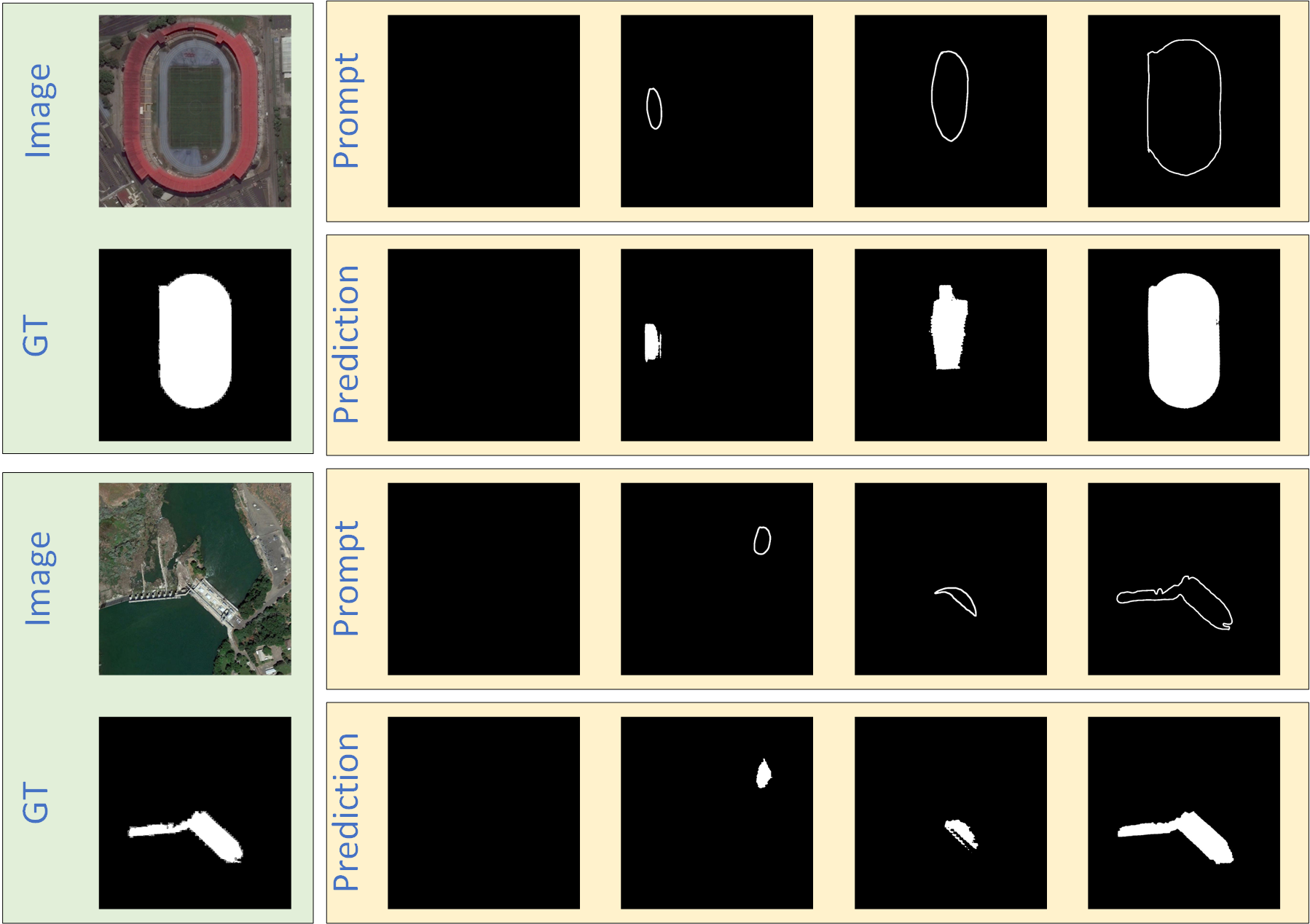}
    \caption{The Impact of Sketch Guidance on the Model}
    \label{fig:5}

\end{figure}
\begin{table*}[htbp]
\centering
\caption{Our method achieves state-of-the-art results in remote sensing image segmentation benchmarks; the best results are highlighted in bold.}
\vskip -0.1in
\label{tab:performance_comparison}
\adjustbox{max width=\textwidth}{
\begin{tabular}{l|c|c|c|c|c|c|c|c|c|c|c|c|c|c|c}
\hline
\multirow{2}{*}{Method}  & \multirow{2}{*}{Prompt} & \multicolumn{2}{c|}{P@0.5 (\%)} & \multicolumn{2}{c|}{P@0.6 (\%)} & \multicolumn{2}{c|}{P@0.7 (\%)} & \multicolumn{2}{c|}{P@0.8 (\%)} & \multicolumn{2}{c|}{P@0.9 (\%)} & \multicolumn{2}{c|}{oIoU (\%)} & \multicolumn{2}{c}{mIoU (\%)} \\
\cline{3-4} \cline{5-6} \cline{7-8} \cline{9-10} \cline{11-12} \cline{13-14} \cline{15-16}
  & & Val & Test & Val & Test & Val & Test & Val & Test & Val & Test & Val & Test & Val & Test\\
\hline
\multirow{5}{*}{SAM \cite{kirillov2023segment}}
& Internal Point  & 55.11 & 56.94  & 46.78 & 48.00  & 35.06 & 35.22  & 23.16 & 22.15  & 10.34 & 10.23 & 36.64 & 38.56 & 50.25 & 51.79 \\
& 1-Point  & 28.45 & 19.13 & 23.51 & 15.20 & 16.26 & 10.26 & 9.48 & 5.49 & 2.99 & 1.72 & 18.66 & 13.21 & 28.47 & 21.55    \\
& 10-Point  & 53.74 & 43.69 & 42.59 & 33.55 & 29.31 & 22.38 & 14.94 & 11.72 & 3.10 & 3.73 & 52.16 & 46.71 & 50.52 & 44.51   \\
& 50-Point  & 39.83 & 34.96 & 26.90 & 22.95 & 13.68 & 12.32 & 3.62 & 4.45 & 0.17 & 0.43 & 50.80 & 47.14 & 41.53 & 38.64    \\
& 100-Point  & 23.68 & 21.78 & 12.13 & 12.09 & 4.31 & 5.14 & 0.69 & 1.01 & 0.06 & 0.06 & 40.70 & 38.21 & 32.64 & 30.63  \\
& 1000-Point  & 0.75 & 0.78 & 0.29 & 0.14 & 0.00 & 0.00 & 0.00 & 0.00 & 0.00 & 0.00 & 14.09 & 12.03 & 13.68 & 9.29 \\
% & Box  & 90.92 & 91.78 & 81.26 & 82.02 & 67.41 & 68.28 & 53.16 & 51.25 & 29.43 & 28.41 & 77.14 & 78.17 & 76.13 & 76.31 \\
%& Sketch(No Train)  & 39.83 & 34.96 & 26.90 & 22.95 & 13.68 & 12.32 & 3.62 & 4.45 & 0.17 & 0.43 & 50.80 & 47.14 & 41.53 & 38.64    \\
% & Sketch  & 69.67 & 61.25 & 62.52 & 52.77 & 51.54 & 42.06 & 41.68 & 28.93 & 23.30 & 13.73 & 71.80 & 68.34 & 60.88 & 53.29 \\
% \hline
% \multirow{2}{*}{SAM-HQ \cite{ke2023segment}}
% & Internal Point  & 51.44 & 53.29 & 40.63 & 42.17 & 28.91 & 29.30 & 17.59 & 17.72 & 7.59 & 7.07 & 37.78 & 39.82 & 47.76 & 49.52  \\
% & 50-Point  & 39.77 & 32.12 & 23.68 & 18.79 & 10.80 & 9.37 & 3.05 & 2.64 & 0.17 & 0.20 & 45.20 & 41.14 & 42.48 & 39.01 \\
\hline
SAM-HQ \cite{ke2023segment} & Internal Point  & 51.44 & 53.29 & 40.63 & 42.17 & 28.91 & 29.30 & 17.59 & 17.72 & 7.59 & 7.07 & 37.78 & 39.82 & 47.76 & 49.52 \\
%& Sketch(No Train)  & 39.77 & 32.12 & 23.68 & 18.79 & 10.80 & 9.37 & 3.05 & 2.64 & 0.17 & 0.20 & 45.20 & 41.14 & 42.48 & 39.01 \\
\hline
\hline
LGCE~\cite{yuan2024rrsis} & Text  & 68.10 & 67.65 & 60.52 & 61.53 & 52.24 & 51.45 & 42.24 & 39.62 & 23.85 & 23.33 & 76.68 & 76.34 & 60.16 & 59.37 \\
\hline
LAVT~\cite{yang2022lavt} & Text  & 69.54 & 69.52 & 63.51 & 63.63 & 53.16 & 53.29 & 43.97 & 41.60 & 24.25 & 24.94 & 77.59 & 77.19 & 61.46 & 61.04 \\
\hline
RMSIN~\cite{liu2024rotated} & Text  & 74.66 & 74.26 & 68.22 & 67.25 & 57.41 & 55.93 & 45.29 & 42.55 & 24.43 & 24.53 & 78.27 & 77.79 & 65.10 & 64.20 \\
\hline
\multirow{2}{*}{SAM* \cite{kirillov2023segment}} & Text  & - & 59.77 & - & 54.31 & - & 46.78 & - & 37.58 & - & 22.24 & - & - & - & - \\
 & Sketch  & 69.67 & 61.25 & 62.52 & 52.77 & 51.54 & 42.06 & 41.68 & 28.93 & 23.30 & 13.73 & 71.80 & 68.34 & 60.88 & 53.29 \\
\hline
UniRef++~\cite{wu2023uniref++} & Sketch  & 87.92 & 86.12 & 83.72 & 80.01 & 72.50 & 67.11 & 57.09 & 49.27 & 34.65 & 26.31 & 85.77 & 84.14 & 75.10 & 71.89 \\
\hline
\textbf{Ours} & Sketch & \textbf{97.04} & \textbf{96.21} & \textbf{92.48} & \textbf{89.57} & \textbf{80.76} & \textbf{75.84} & \textbf{66.58} & \textbf{57.40} & \textbf{40.69} & \textbf{31.46} & \textbf{90.71} & \textbf{88.49} & \textbf{82.24} & \textbf{79.58} \\
\hline
\end{tabular}}
\vspace{-0.1cm}
\label{tab:1}
\end{table*}

%\subsection{Comparison with state-of-the-art RIS methods}

%In our experiments, we compared the performance of our method with existing state-of-the-art image segmentation methods on the validation and test sets of the RRSIS-D dataset (see Table \ref{tab:1}). To ensure a fair comparison, we used the original implementation details of these comparative models. It is worth noting that our method significantly outperforms the other methods on both subsets. Most notably, the mIoU is 6.11\% and 3.27\% higher than the closest SAM on the validation and test subsets, respectively. This leap in performance is particularly evident in complex scenarios, such as detecting large targets with complex and irregular shapes or rotated objects (see Fig. \ref{fig:4}).
\begin{figure*}[htbp] 
    \centering 
    \includegraphics[width=0.9\textwidth]{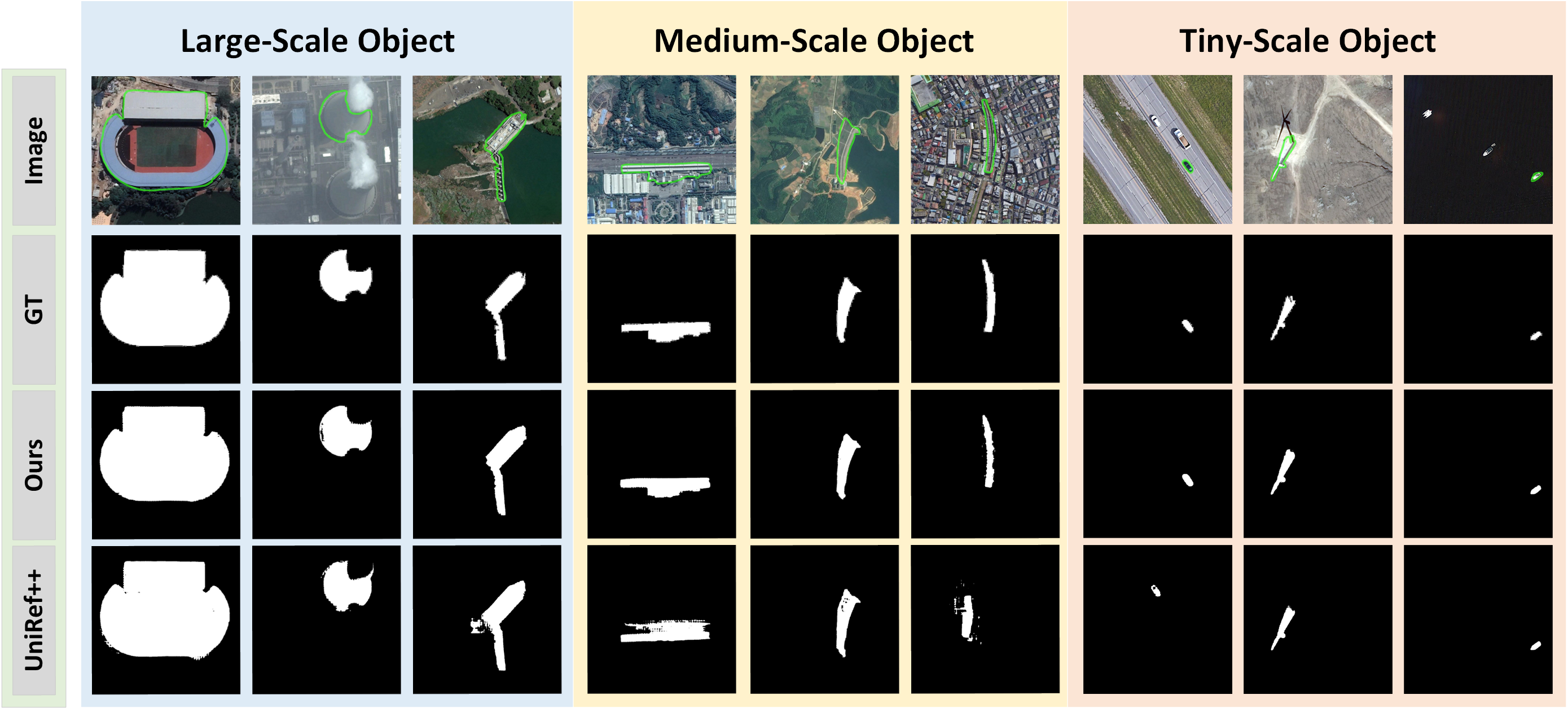} 
    \caption{Our model can effectively handle predictions of objects with different sizes, surpassing existing approaches.}
    \label{fig:4} 
    \vspace{-0.5cm}
\end{figure*}
\subsection{Ablation Study}
\noindent\textbf{Sketch Guidance.}
% \subsubsection{Sketch Guidance.}
To verify that the model performs segmentation under the guidance of sketches, we designed a set of experiments as shown in Fig. \ref{fig:5}. When no information is drawn, the network does not produce any output. When a portion of the area is drawn, the network accurately outputs results for the different regions. It is also evident that the network does not simply fill in the sketch; rather, it selects meaningful areas for output based on the actual shapes of the objects in the image.

\noindent\textbf{The Better the Drawing, The Better the Result.}
% \subsubsection{The Better the Drawing, The Better the Result}
We made full use of the sketch augmentation module to apply varying degrees of perturbation to the input sketches, simulating the sketch deviations from hands and differences in drawing skills. We set the perturbation intensity \( K \) to 10 as Low Pertubation and 25 as High Pertubation. As shown in  Table \ref{tab:2}, less errors in the test set, better the result (which makes intuitive sense).

% \begin{table*}[htbp]
% \centering
% \caption{Perturbations of varying degrees applied to input sketches via the sketch augmentation module simulate deviations in sketches during the actual hand-drawing process, and the experimental results detailedly document the performance of the model.}
% \renewcommand{\arraystretch}{1.2}  % 调整行间距
% \adjustbox{max width=\textwidth}{
% \begin{tabular}{l|c|c|c|c|c|c|c|c}
% \hline
% \multirow{1}{*}{Method}  & \multirow{1}{*}{perturbation intensity} & \multicolumn{1}{c|}{P@0.5 (\%)} & \multicolumn{1}{c|}{P@0.6 (\%)} & \multicolumn{1}{c|}{P@0.7 (\%)} & \multicolumn{1}{c|}{P@0.8 (\%)} & \multicolumn{1}{c|}{P@0.9 (\%)} & \multicolumn{1}{c|}{oIoU (\%)} & \multicolumn{1}{c}{mIoU (\%)} \\
% \cline{3-9}
% \hline
% \multirow{4}{*}{\textbf{Ours}}
% & High   & 56.94  & 47.06  & 34.04  & 20.11  & 5.66  & 72.57  & 49.36 \\
% & Medium  & 73.23   & 63.95   & 50.42   & 33.55  & 12.55  & 80.74  & 61.85  \\
% & Low   & 91.41  & 83.74  & 69.61  & 51.88  & 25.80  & 87.04  & 76.04 \\
% & Normal  & \textbf{96.21}  & \textbf{89.57}  & \textbf{75.84}  & \textbf{57.40}  & \textbf{31.46}  & \textbf{88.49}  & \textbf{79.58} \\
% \hline
% \end{tabular}}
% \label{tab:2}
% \end{table*}

\begin{table}[H]
\centering
\caption{Perturbations of varying degrees applied to input sketches in LTL-Sensing dataset via the sketch augmentation module}
\renewcommand{\arraystretch}{1.3}  % 调整行间距
\footnotesize  % 增大字体大小
\resizebox{\columnwidth}{!}{  % 调整表格宽度以适应单栏
\begin{tabular}{l|ccccccc}
\hline
Perturbation Intensity & P@0.5 & P@0.6 & P@0.7 & P@0.8 & P@0.9 & oIoU & mIoU \\
\hline
High Perturbation  & 73.23 & 63.95  & 50.42  & 33.55  & 12.55  & 80.74  & 61.85  \\
Low Perturbation   & 91.41 & 83.74  & 69.61  & 51.88 & 25.80  & 87.04  & 76.04 \\
---  & \textbf{96.21} & \textbf{89.57} & \textbf{75.84} & \textbf{57.40} & \textbf{31.46}  & \textbf{88.49}  & \textbf{79.58} \\
\hline
\end{tabular}
}
\label{tab:2}
\end{table}

%From the experimental results, it is evident that despite the noticeable irregular changes and variations in the sketches due to perturbations, the segmentation results output by the model were not significantly adversely affected. This phenomenon strongly indicates that our model possesses a high degree of flexibility and can tolerate, to some extent, the quality deviations in sketch inputs. In other words, when guided by sketches, the model's dependence on sketch quality is relatively low. Even if the sketches are not perfect, the model can still rely on its own capabilities to use the key information conveyed by the sketches to complete relatively accurate segmentation tasks. This fully demonstrates the reliability and practicality of the model in real-world applications.

\noindent\textbf{Impact of the Fusion Mechanism.}
% \subsubsection{Impact of Masked Attention and MPT}
In Tab. \ref{tab:uniref}, we show that adding the fusion mechanism can effectively boost the performance of the network. We add the fusion mechanism to SAM (w/ fusion mechanism, use the vanilla attention) while keep other configuration the same.  

\begin{table}[H]
\caption{Effectiveness of Fusion Mechanism}
\vskip -0.3in
\label{sample-table}
\begin{center}
\resizebox{\columnwidth}{!}{ % 调整表格宽度以适应栏宽
\begin{small}
% \begin{sc}
\begin{tabular}{c|ccccc}
\toprule
Model  & P@0.5  & P@0.7  & P@0.9  & oIoU  & mIoU  \\
\midrule
SAM  &61.25   &42.06   &13.73  &68.34  &53.29 \\
% SAM w/ fusion mechanism &86.12  & 67.11  &26.31 &84.14 &71.89 \\
 \textbf{SAM w/ fusion mechanism} &  \textbf{86.12}  &  \textbf{67.11}  &  \textbf{26.31} &  \textbf{84.14} &  \textbf{71.89} \\
% $\times$    &$\surd$   &55.62 &36.00  &9.94 &63.13 &48.53\\
% Ours  & \textbf{96.21}  & \textbf{75.84} & \textbf{31.46} & \textbf{88.49} & \textbf{79.58}       \\
\bottomrule
\end{tabular}
% \end{sc}
\end{small}
}
\end{center}

\vspace{-0.5cm}
\label{tab:uniref}
\end{table}

\noindent\textbf{Impact of MPT and Masked Attention.}
In Table \ref{tab:3}, we show that that the introduction of MPT and masked attention significantly enhances the model's performance. For more ablation studies and details, please refer to the supplementary material. 

\begin{table}[H]
\caption{Effectiveness of Masked Attention and MPT (Multi-Prompts Transport).}

\label{sample-table}
\begin{center}
\resizebox{\columnwidth}{!}{ % 调整表格宽度以适应栏宽
\begin{small}
% \begin{sc}
\begin{tabular}{c|ccccc}
\toprule
Model  & P@0.5  & P@0.7  & P@0.9  & oIoU  & mIoU  \\
\midrule
w/o Masked Attention, w/o MPT  &86.12   &67.11   &26.31  &84.14  &71.89 \\
w/o MPT &95.98  & \textbf{76.76}  &31.28 &87.08 &79.48 \\
% $\times$    &$\surd$   &55.62 &36.00  &9.94 &63.13 &48.53\\
Ours  & \textbf{96.21}  & 75.84 & \textbf{31.46} & \textbf{88.49} & \textbf{79.58}       \\
\bottomrule
\end{tabular}
% \end{sc}
\end{small}
}
\end{center}
\vskip -0.1in
\label{tab:3}
\end{table}

% \begin{table}[htbp]
% \centering
% \caption{--}
% \renewcommand{\arraystretch}{!}  % 调整行间距

% \begin{tabular}{c|c|c|c|c}
% \hline
% Method & Sketch-AssistedSaliency Detection & Sketch-basedSegtmentation & InteractiveSegmen-tationFoundationModel & Ours &  \\
% \hline
% Spatial Specifcation  &  &   & $\surd$  & $\surd$    \\
% \hline
% Human-guided lmprovement   & $\surd$  & $\surd$  & $\surd$  & $\surd$  \\
% \hline
% Input Error-Tolerant Flexibility   &   & $\surd$  &  & $\surd$  \\
% \hline
% \end{tabular}

% \label{tab:2}
% \end{table}

%-------------------------------------------------------------------------
\section{Conclusion}
% This work enhances zero-shot interactive segmentation for remote sensing imagery through three pivotal contributions. First, we introduced a novel sketch-based prompting approach that empowers users to intuitively outline objects, overcoming the limitations of traditional point or bounding box prompts. Second, we developed LTL-Sensing, the first dataset that combines human freehand sketches with remote sensing imagery, establishing a valuable benchmark for future research. Finally, we presented LTL-Net, a dedicated model that incorporates a multi-input prompting transport module to effectively manage the unique characteristics of freehand sketches. Our extensive experiments demonstrate that this approach markedly increases segmentation accuracy and robustness compared to state-of-the-art methods like SAM. This progress paves the way for more intuitive and efficient human-AI collaboration in remote sensing analysis, ultimately enhancing the effectiveness of various applications in this critical field.
This work advances zero-shot interactive segmentation for remote sensing imagery through three key contributions. First, we propose a novel sketch-based prompting method, enabling users to intuitively outline objects with greater precision, surpassing traditional point or box prompts that often struggle with complex or ambiguous boundaries. Second, we introduce LTL-Sensing, the first dataset pairing human sketches with remote sensing imagery, providing a high-quality benchmark that facilitates future research and fosters more effective human-in-the-loop segmentation techniques. Third, we present LTL-Net, a model featuring a multi-input prompting transport module specifically designed to process freehand sketches, ensuring seamless integration with existing segmentation frameworks. Extensive experiments demonstrate that our approach significantly enhances segmentation accuracy and robustness compared to state-of-the-art methods like SAM, paving the way for more intuitive human-AI collaboration in remote sensing analysis and expanding its real-world applications in areas such as environmental monitoring and disaster response.
%-------------------------------------------------------------------------

{
    \small
    \clearpage
    \bibliographystyle{ieeenat_fullname}
    \bibliography{main}
}
\clearpage
\setcounter{page}{1}
\maketitlesupplementary
\setcounter{table}{0}
\setcounter{figure}{0}
\setcounter{section}{0}
\setcounter{equation}{0}

\section{Impact of Mask Size}
\begin{figure}[h]
    \centering
    \includegraphics[width=\columnwidth]{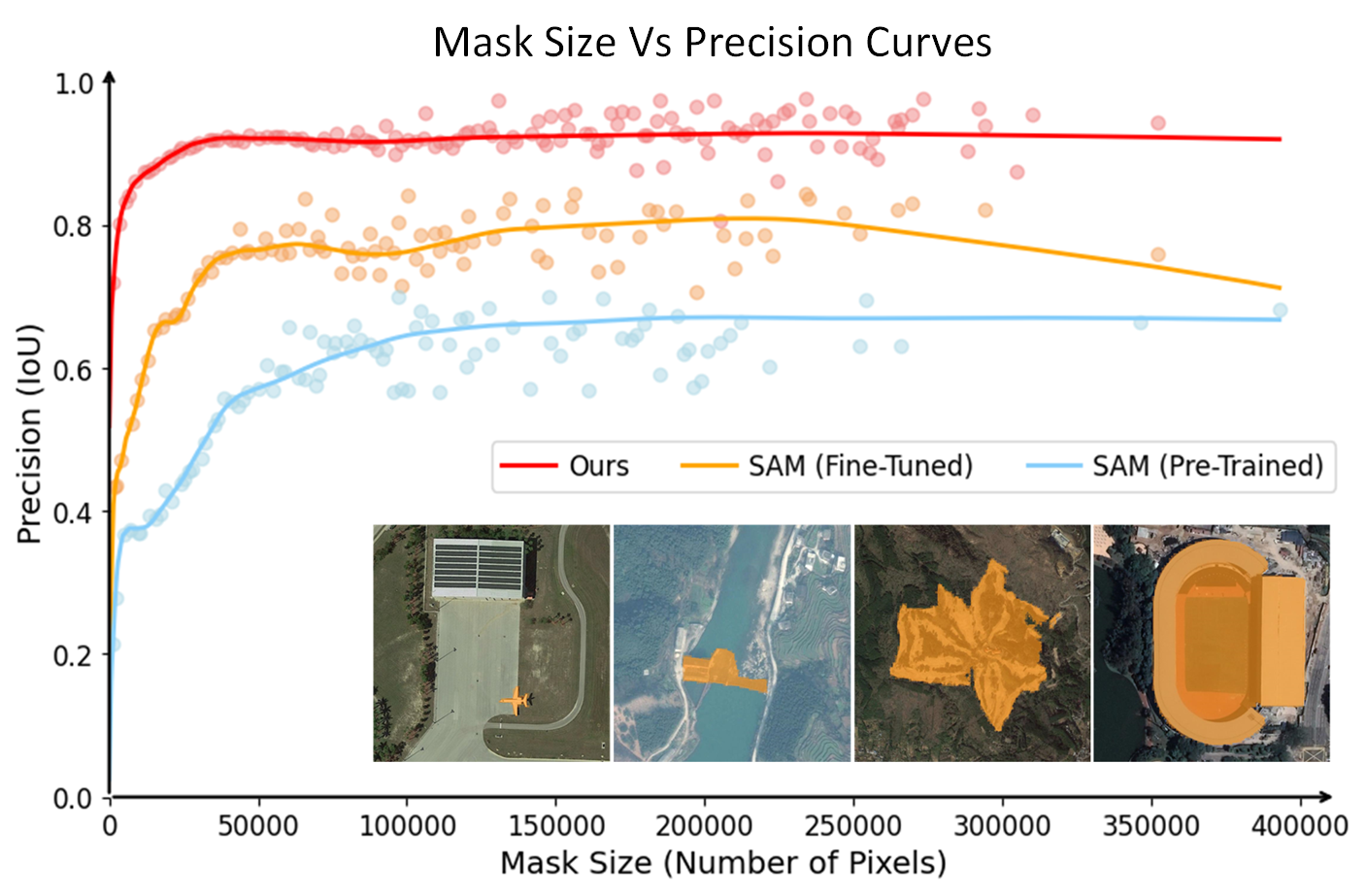}
    \caption{Our carefully designed LTL-Net is capable of producing more accurate segmentation masks in all mask size ranges with sketch input compared to vanilla SAM (Pre-Trained) and SAM fine-tuned on remote sensing image dataset with same input prompts, demoted as SAM (Fine-Tuned). More details in Supplementary Material.}
    \label{fig:x2}
\end{figure}

    To investigate the performance differences of the model on targets of different scales, we designed the following experiments. Specifically, it should be noted that the SAM(fine-tuned) in the images uses sketch input, while the SAM(pre-trained) employs point-sampling to simulate sketch input, with a point-sampling number of 50. We used the number of pixels occupied by the target in the ground-truth mask as a quantitative measure of target size and calculated the IoU value between it and the predicted mask on a per-sample basis. Based on the full data statistics of the test set, we employed locally weighted regression to locally fit the scatter plot of target size and IoU, which can effectively capture the local trends in the data. As shown in Fig. \ref{fig:x2}.

    The algorithm utilizes \textbf{LOWESS} curve fitting with a smoothing coefficient \texttt{frac=0.2}, determined through grid search to optimally balance detail preservation and noise suppression. Adaptive \textbf{IoU thresholds} (e.g., $\text{IoU} \geq 0.8$ for Ours model in regions beyond 50,000 pixels) are set based on training set statistics. For data sampling, a fine-grained binning strategy (\texttt{bin\_size=2000}) is applied to dense regions below 50,000 pixels, selecting \texttt{1} closest point per bin (\texttt{samples\_per\_bin=1}), while higher pixel ranges adopt sparser sampling to avoid overplotting. 
    
    The results of the fitted curve indicate that our model outperforms the SAM model across different target sizes. Specifically, our model demonstrates higher accuracy and stability, whether dealing with small or large targets. This advantage is particularly evident when handling large targets, where our model exhibits more stable performance and a significant improvement compared to the SAM model. This superiority is especially prominent in the segmentation tasks of large-scale targets, indicating that our model can better maintain the accuracy and consistency of segmentation for large targets.

\section{LTL-Sensing Dataset visualization}
\begin{figure}[H]
    \centering
    \includegraphics[width=\columnwidth]{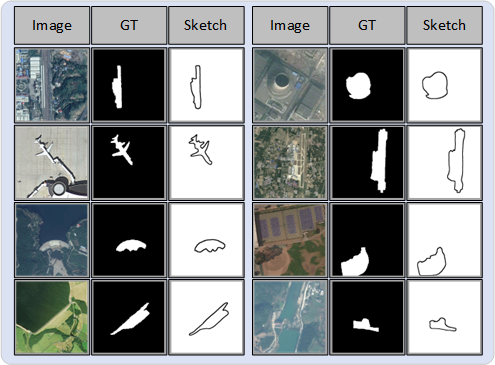}
    \caption{Partial sample display of LTL-Sensing Dataset}
    \label{fig:x1}
\end{figure}

    In the supplementary materials of this paper, we present sample visualizations from the LTL Sensing dataset (Fig. \ref{fig:x1}). Each triplet contains: the original remote sensing image, professionally annotated semantic segmentation ground-truth (GT), and the corresponding hand-drawn sketch.
    
    This collection of hand-drawn sketches helps researchers study how people understand visual patterns and translate them into symbols. The dataset contains detailed city scenes, demonstrating that our approach works effectively across different scales - whether analyzing small objects or large urban spaces, and regardless of how simple or complex the subject matter is.

\section{Algorithm of Sketch Augmentation}
    The Sketch Augmentation employs structure-aware perturbation through hierarchical block processing and Bézier curve deformation, where control point displacements are adaptively scaled based on local sketch density. This approach preserves geometric integrity while generating diverse variants, as detailed in Algorithm \ref{alg:bezier_aug}.
    
    To process the hand-drawn sketches, we first convert them to grayscale images and apply binarization with a threshold of \( t = 128 \) to extract the skeleton structure of the sketches. Next, we use a morphological skeletonization method to obtain the centerlines \( S_0 \), which preserves the topological structure of the image while removing boundary pixels. We then segment the centerlines \( S_0 \) into non-overlapping square blocks of size \( 32 \times 32 \), suitable for sketches of size \( 256 \times 256 \). Considering the thickness of sketch lines, we extract the centerlines before segmentation to prevent lines from being split into multiple segments. Within each block, we select the largest set of connected pixels as the main curve \( T \) and fit it using cubic Bezier curves. The displacement increment \( K \) used in this process is set to 10.

\begin{algorithm}[htbp]
\caption{Sketch Augmentation using Bezier Curve Fitting}
\label{alg:bezier_aug}

\begin{algorithmic}[1]
\Require Original sketch $S$, perturbation control parameters $C$ and $K$
\Ensure Augmented sketch set $\{S'_1, S'_2, ..., S'_K\}$

\State $S_{\text{bin}} \gets \text{Binarize}(S, \text{threshold}=128)$
\State $S_{\text{skel}} \gets \text{Skeletonize}(S_{\text{bin}}) \triangleright \text{Zhang-Suen algorithm}$

\State $\text{Blocks} \gets \text{DivideIntoBlocks}(S_{\text{skel}}, \text{block\_size}=200)$

\For{each block $B \in \text{Blocks}$}
    \State $\text{CC} \gets \text{ConnectedComponents}(B)$
    \For{each component $c \in \text{CC}$ with $|c| \geq 10$ pixels}
\State $P \gets \text{SamplePoints}(c, \text{interval}=10)$ \Comment{Sample every 10 pixels}
\State $\{p_0, p_1, p_2, p_3\} \gets \text{FitBezierCurve}(P)$ \Comment{Fit cubic Bezier curve}
\State $\text{row} \gets \text{GetRowCount}(c)$
\State $\theta \gets \left\lfloor \frac{\text{row}}{C} \right\rfloor \times K$ \Comment{Perturbation magnitude}

\For{$i = 1$ \textbf{to} $K$}
    \State $\delta_1, \delta_2 \gets \mathcal{N}(0, \theta)$ \Comment{2D Gaussian noise}
    \State $p'_1 \gets p_1 + \delta_1$ \Comment{Perturb control points}
    \State $p'_2 \gets p_2 + \delta_2$
    \State $B'_i \gets \text{RenderBezier}(p_0, p'_1, p'_2, p_3)$
\EndFor
    \EndFor
\EndFor

\State Combine perturbed blocks to generate $K$ augmented sketches
\State \Return $\{S'_1, S'_2, ..., S'_K\}$
\end{algorithmic}
\end{algorithm}
% \vskip -0.5in

\section{Algorithm of Multi-Prompts Transport (MPT)}
The Multi-Prompts Transport (MPT) framework establishes dense cross-modal alignment through entropy-regularized optimal transport, adaptively fusing heterogeneous sketch guidance while preserving spatial coherence. As shown in Algorithm \ref{alg:mpt}, this process operates in three coordinated stages: (1) multi-scale similarity computation, (2) Sinkhorn-regularized transport optimization, and (3) geometry-aware score aggregation. Critical implementation parameters include the marginal distributions $\mu \in \mathbb{R}^n_+$ and $\nu \in \mathbb{R}^m_+$ (initialized as uniform priors $\mu = \frac{1}{n}\mathbf{1}_n$ and $\nu = \frac{1}{m}\mathbf{1}_m$ for $n$ sketch prompts and $m$ image regions), the entropy regularization coefficient $\varepsilon = 0.05$ controlling transport plan smoothness, and Sinkhorn convergence criteria (\texttt{max\_iter}=50, \texttt{tolerance}=$10^{-4}$). Spatial dimensions $H \times W$ correspond to the input image resolution, while $K = n$ explicitly denotes the number of sketch prompts, enabling channel-wise fusion through the \texttt{Reshape}([H, W, K]) operation.

\begin{algorithm}[htbp]
\caption{Multi-Prompts Transport (MPT)}
\label{alg:mpt}
\begin{algorithmic}[1]
\Require Image features $\mathbf{F}_i$, sketch features $\{\mathbf{F}_{s_1}, \mathbf{F}_{s_2}, ..., \mathbf{F}_{s_n}\}$
\Ensure Optimized score map $\mathbf{S}^*$

\State Initialize $\mathbf{b}^0 \gets \mathbf{0}$ \Comment{Log-domain dual variables}
\For{each sketch feature $\mathbf{F}_{s_j}$}
    \State $\mathbf{S}_j \gets \mathbf{F}_{s_j} \mathbf{F}_i^\top$ \Comment{Cosine similarity matrix}
\EndFor
\State $\mathbf{S} \gets \mathrm{Concat}([\mathbf{S}_1, \mathbf{S}_2, ..., \mathbf{S}_n])$ \Comment{Stack score maps}
\State $\mathbf{C} \gets \mathbf{1} - \mathbf{S}$ \Comment{Cost matrix conversion}

\For{$t = 1$ \textbf{to} $\mathrm{max\_iter}$} \Comment{Sinkhorn iteration}
    \State $\mathbf{a}^t \gets \log(\mu) - \mathrm{LogSumExp}(-\mathbf{C}/\varepsilon + \mathbf{b}^{t-1})$
    \State $\mathbf{b}^t \gets \log(\nu) - \mathrm{LogSumExp}(-\mathbf{C}/\varepsilon + \mathbf{a}^t)$
    \If{$\|\mathbf{b}^t - \mathbf{b}^{t-1}\|_2 < \mathrm{tolerance}$}
        \State \textbf{break}
    \EndIf
\EndFor

\State $\mathbf{T}^* \gets \exp(\mathbf{a}^t) \odot \exp(-\mathbf{C}/\varepsilon) \odot \exp(\mathbf{b}^t)$ \Comment{Optimal transport plan}
\State $\mathbf{S}^* \gets \mathrm{Reshape}(\mathbf{T}^* \odot \mathbf{S}, [H, W, K])$ \Comment{Transport application}
\State $\mathbf{S}^* \gets \mathrm{Sum}(\mathbf{S}^*, \mathrm{dim}=3)$ \Comment{Aggregation}
\State \Return $\mathbf{S}^*$
\end{algorithmic}
\end{algorithm}

\section{Other Evaluation Metrics}
    
    To comprehensively evaluate the performance of the proposed model, we adopt other four widely-used metrics in saliency detection/segmentation tasks (Tab. \ref{tab:performance_comparison}), following the evaluation protocol in \cite{chen2023sam}:

\begin{table}[h]
\centering
\caption{Performance Comparison of Different Methods}
\label{tab:performance_comparison}
\adjustbox{max width=\columnwidth}{
\begin{tabular}{lccccc}
\hline
\multirow{2}{*}{Method}  & \multirow{2}{*}{Prompt} & \multicolumn{4}{c}{Test Dataset}  \\
\cline{3-6} 
  & & $S_m\uparrow$ & $E_m \uparrow$ & $F^{w}_m \uparrow$ & $\text{MAE} \downarrow$ \\
\hline
\multirow{6}{*}{SAM}
& \multicolumn{1}{l}{Internal-Point}  & 0.7534 & 0.7879 & 0.6099 & 0.0399   \\
& \multicolumn{1}{l}{1-Point}  & 0.5556 & 0.5784 & 0.2516 & 0.1053    \\
& \multicolumn{1}{l}{10-Point}  & 0.7054 & 0.7379 & 0.4807 & 0.0494    \\
& \multicolumn{1}{l}{100-Point}  & 0.6087 & 0.6696 & 0.3394 & 0.0613   \\
& \multicolumn{1}{l}{1000-Point}  & 0.4386 & 0.4986 & 0.1110 & 0.1665   \\
% & \multicolumn{1}{l|}{Box}  & 0.8872 & 0.9721 & 0.8691 & 0.0107 & 0.8854 & 0.9687 & 0.8654 & 0.0116   \\
& \multicolumn{1}{l}{Sketch}  & 0.7610 & 0.9160 & 0.6129 & 0.0162    \\
\hline
\multirow{2}{*}{SAM HQ}
& \multicolumn{1}{l}{Internal-Point}  & 0.7429 & 0.7997 & 0.6055 & 0.0337    \\
% & \multicolumn{1}{l|}{Box}  & 0.8726 & 0.9697 & 0.8510 & 0.0091 & 0.8709 & 0.9658 & 0.8473 & 0.0101   \\
& \multicolumn{1}{l}{Sketch}  & 0.6638 & 0.7121 & 0.4336 & 0.0639    \\
\hline
\multirow{1}{*}{UniRef++}
& \multicolumn{1}{l}{Sketch}  & 0.8701 & 0.9720 & 0.7920 & 0.0080   \\
% & \multicolumn{1}{l|}{Edge}  & 0.8717 & 0.9727 & 0.7960 & 0.0084 & 0.8856 & 0.9758 & 0.8193 & 0.0082   \\
\hline
\multirow{1}{*}{\textbf{Ours}}
& \multicolumn{1}{l}{Sketch} & \textbf{0.9124} & \textbf{0.9847} & \textbf{0.8702} & \textbf{0.0057}  \\
% & \multicolumn{1}{l|}{Edge} & \textbf{0.9149} & \textbf{0.9855} & \textbf{0.8749} & \textbf{0.0058} & \textbf{0.9248} & \textbf{0.9885} & \textbf{0.8905} & \textbf{0.0052} \\
\hline
\end{tabular}}
\end{table}

    Structure-measure ($S_m$): Quantifies the structural similarity between predicted maps and ground truth, emphasizing both region-aware and object-aware structural consistency.
    
    Enhanced-alignment-measure ($E_m$): Evaluates both pixel-level matching and image-level statistics by combining local and global similarity through adaptive thresholding.
    
    Weighted F-measure ($F_\beta^w$): Improves traditional F-measure by introducing weights based on pixel position importance, providing a balanced assessment of precision and recall.
    
    Mean Absolute Error (MAE): Computes the pixel-wise average absolute difference between predictions and ground truth, directly reflecting overall error magnitude.
    
    These metrics complement each other: ($S_m$) and ($E_m$) focus on structural and edge alignment, ($F_\beta^w$) evaluates region-level consistency with positional awareness, while MAE provides straightforward pixel-level accuracy measurement. The combination ensures a holistic evaluation from both statistical and visual perspectives.

\section{More Experimental Details}
\noindent\textbf{Curation of the Dataset}
The training data for the sketch input is derived from Canny edge detector parsing the RRSIS-D dataset (GT mask). The data curation protocol follows \cite{zang2025let}.

\noindent\textbf{Why Not Fine-Tune SAM with Point/Box Prompt?}
Notably, since the RRSIS-D dataset itself originates from SAM’s pretrained network (Pixel-level masks for all images in the dataset are generated using SAM, leveraging bounding box prompts sourced from the RSVGD Dataset), further fine-tuning on this dataset alone would be redundant unless additional feature encoders are introduced. To explore this, we fine-tuned SAM with two different encoders: using a CLIP encoder to incorporate textual input and using our ResNet encoder to introduce sketch-based guidance. Moreover, measuring the box prompt for SAM is also meaningless because it is where GT comes from. In our main experiment, SAM and SAM-HQ are pre-trained, other method, including SAM* are fine-tuned / trained with RRSIS-D dataset. The configuration for fine-tuning SAM with text input is from \cite{chen2025rsrefseg}.

\end{document}